\documentclass[letterpaper, 10 pt, conference]{ieeeconf}
\usepackage{amsmath,amsfonts}
\usepackage{graphicx}

\usepackage{url}
\PassOptionsToPackage{hyphens}{url}\usepackage{hyperref}

\usepackage{booktabs}
\usepackage{multirow}
\usepackage[noadjust]{cite}

\usepackage{xcolor}

\usepackage{pifont}
\usepackage{newunicodechar}
\newunicodechar{✓}{\ding{51}}
\newunicodechar{✗}{\ding{55}}

\newcommand\parag[1]{\smallskip\noindent {\bf #1.}}

\IEEEoverridecommandlockouts                              

\author{Lina Mezghani$^{1, 2}$ , Sainbayar Sukhbaatar$^{1}$, Thibaut Lavril$^{1}$, Oleksandr Maksymets$^{1}$, \\ Dhruv Batra$^{1, 3}$, Piotr Bojanowski$^{1}$, Karteek Alahari$^{2}$
\thanks{$^{1}$ Meta AI}%
\thanks{$^{2}$ Univ.\ Grenoble Alpes, Inria, CNRS, Grenoble INP, LJK, 38000 Grenoble, France}%
\thanks{$^{3}$ Georgia Institute of Technology}%
   }

\title{\LARGE \bf
Memory-Augmented Reinforcement Learning for Image-Goal Navigation
}

\begin{document}

\maketitle
\thispagestyle{empty}
\pagestyle{empty}

%

\begin{abstract}
In this work, we present a memory-augmented approach for image-goal navigation.
Earlier attempts, including RL-based and SLAM-based approaches have either shown poor generalization performance, or are heavily-reliant on pose/depth sensors.
Our method is based on an attention-based end-to-end model that leverages an episodic memory to learn to navigate.
First, we train a state-embedding network in a self-supervised fashion, and then use it to embed previously-visited states into the agent’s memory.
Our navigation policy takes advantage of this information through an attention mechanism.
We validate our approach with extensive evaluations, and show that our model establishes a new state of the art on the challenging Gibson dataset.
Furthermore, we achieve this impressive performance from RGB input alone, without access to additional information such as position or depth, in stark contrast to related work.
\end{abstract}

\section{Introduction}

The challenges of addressing navigation tasks go beyond the classical computer-vision setup of learning from pre-defined fixed datasets.
They consist of problems such as low-level control point-goal navigation~\cite{wijmans2019dd}, object-goal navigation~\cite{batra2020objectnav} or even tasks requiring natural language understanding, e.g., embodied question answering~\cite{das2018embodied}.

\begin{figure}
  \centering
  \includegraphics{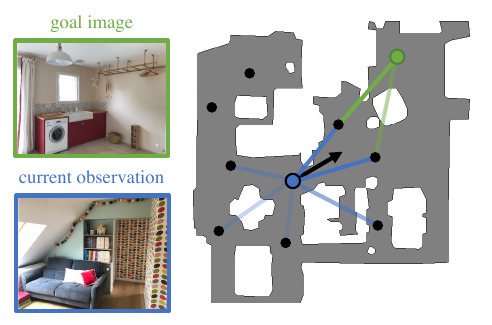}
  \caption{
    We tackle the problem of image-goal navigation.
    The agent (shown as the blue dot) is given an image from a goal location (green dot) which it must navigate to.
    To address this task, our agent stores an episodic memory of visited states (black dots), and uses a navigation policy that puts attention (lines) on this memory (best viewed in pdf).
  }
  \label{fig:pull_figure}
\end{figure}

We focus on one such critical problem: image-goal navigation~\cite{zhu2017target}, wherein an agent has to learn to navigate to a
location specified by visual observations taken from there.
Consider the agent in \autoref{fig:pull_figure}, which is spawned at the location in blue on the map, where it observes a sofa.
The agent's task is to find the location in the house where it would see the washing machine shown in the goal image.
In terms of difficulty, this task lies in between point-goal and object-goal navigation.
Indeed, it does not require learning the association between visual inputs and manual labels (as in object-goal navigation), but it needs a higher-level understanding of scenes for navigating through them.
There are several facets to this task, which make it challenging.

The first challenge is to design methods that are completely end-to-end.
This allows for approaches that require less expert knowledge and are easily transferable to new simulators and tasks.
Classical methods for agent navigation, based on simultaneous localization and mapping~\cite{thrun2002probabilistic}, comprise multiple hand-crafted modules, and require a large amount of annotated data.
Reinforcement learning (RL) is a popular framework for tackling navigation problems in an end-to-end manner~\cite{zhu2017target, fang2019scene}.
However, in the context of photorealistic data, existing methods have either shown results in a limited setting with synthetic data~\cite{fang2019scene} or suffer from poor RL-based performance~\cite{chaplot2020neural}.

The second challenge in image-goal navigation is the need for a high-level understanding of the surrounding scene.
In photorealistic environments, agents trained with RL are subject to overfitting due to the high dimensionality of the data and the limited number of environments available for training~\cite{xia2018gibson}.
Learning to navigate in such visually complex environments, therefore, requires learning a more informative representation than pixels, which captures the visual diversity of the scene and is generalizable to unseen environments.
Previous methods use the position of the agent and/or a depth map~\cite{chaplot2020neural, fang2019scene} to improve the generalization of the model.
Approaches that learn from RGB input only have not shown generalization to unseen environments ~\cite{savinov2018semi, mezghani2020learning}.

Finally, a requirement for navigating agents is to build and exploit a representation of the states visited.
Indeed, the agent should be able to remember the places it has already visited within an episode for efficient exploration~\cite{mezghani2020learning, fang2019scene}.
This ``memory'' can take the form of a buffer~\cite{fang2019scene}, or a metric \cite{thrun2002probabilistic,chaplot2018active} or a topological~\cite{savinov2018semi,chaplot2020neural} map.
In most methods, the agent exploits this information with an explicit planner~\cite{savinov2018semi, chaplot2020neural}, but some work~\cite{fang2019scene} has also leveraged it with an implicit attention.
However, the generalization properties of these schemes remain unclear: the approach by~\cite{savinov2018semi} requires exploration videos collected by a human when navigating to an unseen environment, while \cite{fang2019scene} only shows limited generalization results on synthetic data.

We tackle these three challenges with a memory-augmented reinforcement learning approach.
First, the agent learns representations of the environment in a self-supervised fashion, and acquires a high-level understanding of the surrounding scene.
It then learns a policy for image-goal navigation in an end-to-end manner.
In order to explore and navigate effectively, the policy is conditioned on an external memory module that remembers useful information from the current episode.
With this approach we establish the new state of the art on the challenging image-goal navigation task on the Gibson dataset~\cite{xia2018gibson} by a large margin.
We achieve this performance from RGB input alone, without access to position information, which is a major improvement compared to previous works like~\cite{chaplot2020neural}.

\section{Related Work}
\parag{SLAM-based Methods} The task of navigation has been studied in the context of simultaneous localization and mapping (SLAM) in robotics~\cite{thrun2002probabilistic}.
Several SLAM methods comprise multiple hand-crafted modules to address strictly-defined problems in specific environments~\cite{tomatis2001combining,mur2015orb,choset2001topological}.
These modules have been progressively replaced with learning-based functions: some approaches~\cite{chaplot2018active} implement the localization module with a neural network, while others~\cite{gupta2017cognitive} replace the metric map with a latent mapping of the environment.
Variants of latent mapping also include a topological map whose nodes contain geometric and semantic information about the environment, as well as a global planner that relies on it~\cite{chaplot2020neural}.
Other works replace SLAM entirely by deep models without explicit planning, and instead rely on a map or memory structure~\cite{parisotto2018neural,zhang2017neural,av2019empnet}.
The major drawback of such methods is that they contain multiple modules that are often trained in a supervised fashion, requiring a large amount of annotated data.
Moreover, SLAM-based methods rely on the availability of position~\cite{parisotto2018neural, zhang2017neural} and depth~\cite{chaplot2020neural,av2019empnet} sensors.
In our work, we focus on the realistic and challenging cas with access to RGB input only.

\parag{RL-based Navigation} Another popular class of methods involves training deep models with reinforcement learning to solve navigation tasks without an explicit world representation.
They use end-to-end frameworks with modules that are less hand-crafted than SLAM-based methods, and have shown good performance on synthetic mazes~\cite{mirowski2016learning} as well as real-world data~\cite{mirowski2018learning,chancan2020mvp,habitatsim2real20ral,wijmans2019dd}.
Such methods have also been explored on indoor-scenes datasets, similar to our setup on image-goal~\cite{zhu2017target} and object-goal~\cite{yang2018visual,maksymets2021thda,chaplot2020object} navigation tasks.
They use an actor-critic model whose policy is a function of both the target and the current state.
While episodic RL is a simple and elegant framework for navigation, it ignores the fact that useful information comes from previous episodes.
In our work, we propose to augment the policy used in RL frameworks with an external memory and to attend on it with a novel, dedicated module.

\parag{Combining RL and Planning} A few recent works have augmented RL-based methods with topological structures, like graphs~\cite{wu2019bayesian,beeching2020learning,savinov2018semi,chen2019behavioral,mezghani2020learning} or memory buffers~\cite{fang2019scene,beeching2020egomap,kumar2018visual}.
They store representations of the visited locations and exploit them at navigation time.
The process of building these representations can be done offline~\cite{savinov2018semi,beeching2020learning,mezghani2020learning}, and requires human-generated data in some cases~\cite{savinov2018semi}.
For example, the test phase in~\cite{savinov2018semi} contains a warm-up stage where the agent builds a graph memory from human trajectories.
Alternatives to this manual annotation do exist, such as building a graph directly with reinforcement learning, using the value function of a goal-conditioned policy as edges weights~\cite{eysenbach2019search}, or buffer of past observations~\cite{fang2019scene}.
These methods were evaluated on synthetic datasets, and have not been scalable to high-dimensional visually-realistic setups.
In particular, \cite{fang2019scene} proposes a method related to our model, with a policy that puts attention on an observation buffer.
In contrast, we learn a representation in a self-supervised fashion to store memory efficiently.
We also present results on a large-scale photorealistic dataset, while \cite{fang2019scene} is limited to synthetic setups.

\parag{Exploration and Representation Learning} Closely related to navigation, the task of exploration has also been extensively studied and has led to interesting breakthroughs in representation learning.
In particular, learning to explore unseen environments has been seen through the spectrum of computer vision~\cite{jayaraman2018learning,chaplot2020semantic}, SLAM-based~\cite{chaplot2019learning}, and RL-based~\cite{chen2018learning,devo2020towards} approaches.
Methods such as~\cite{savinov2018episodic,mezghani2020learning} leverage a self-supervised representation learning stage to prepare the exploration phase.
Our work extends this line of study by first showing that a self-supervised pretraining phase allows to learn useful information that generalizes to unseen environments, as well as proposing a novel attention-based navigation policy that takes advantage of this information.
Posterior to our work, there have been extensions of our method, including~\cite{sang2022novel} that exploits a similar approach for object-goal navigation in unseen environments.
This work uses the same components than our method: both short-term and long-term memories, as well as an attention mechanism to leverage past experience.
The main difference with our work relies on the task being tackled: contrary to our work, \cite{sang2022novel} focuses on object-goal navigation, which consists in navigating to an instance of a target object.
\begin{figure*}[t]
  \centering
  \includegraphics[width=0.9\linewidth]{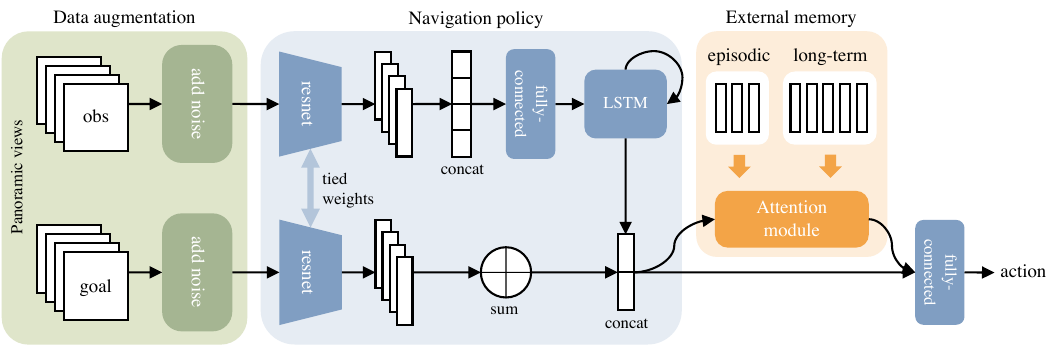}
  \caption{An overview of our model that consists of three parts: a data augmentation module (green) for better generalization, a navigation policy (blue) for picking actions, and an external memory (orange) for conditioning on previous observations.
  }
  \label{fig:model}
\end{figure*}

\section{Problem Formulation}
\label{sec:problem}
We consider the classical formulation of episodic image-goal navigation as defined in~\cite{chaplot2020neural}.
At the beginning of a navigation episode, an agent is given a target observation $x^*$, composed of an RGB image from the target location.
At each timestep $t$, the agent performs an action $a_t$ and receives the next observation $x_{t+1}$ as well as a reward $r_t$ from the environment.
The objective is to learn a navigation policy function $\pi(a_t | x_t, x^*)$ that brings the agent closer to the target location.
We complete the definition of our setup with the following details.

\parag{Action Space} It comprises four actions: \texttt{MOVE\_FORWARD}, \texttt{TURN\_LEFT}, \texttt{TURN\_RIGHT} and \texttt{STOP}. Please refer to \autoref{impl_details} for numerical details.

\parag{Success Criterion} An episode is considered successful if the agent performs the stop action within a range of $l$ from the target location.
In cases where the agent performs the stop action outside of this range, or if the maximum number of steps is exceeded before the agent performs the stop action, the episode is considered a failure.

\parag{Observation Space} The observation of the agent $x_t$ as well as the goal observation $x^*$ are the RGB images of the first-person view at those locations.
Each RGB image is a panoramic sensor of size $v \times 3 \times 128 \times 128$.
We compute this panoramic input by gathering observations from $v$ successive rotations of angle $(360/v)^\circ$ from our agent's location.
Note that we do not have access to neither the agent's position nor any depth sensor information.

\parag{Reward} We follow the classic setup for image-goal navigation~\cite{chaplot2020neural} where the reward is split into three components: 
(i) \textit{sparse success reward}: that rewards the agent for performing the stop action within the success range around the target location, 
(ii) \textit{dense shaping reward}: that is equal to the decrease in distance to the goal,
(iii) \textit{dense slack reward}: that penalizes the agent for being alive at each step, and encourages shorter trajectories.
\section{Our approach}
\label{model}
As shown in \autoref{fig:model}, our agent model has three parts: a data augmentation part for improving generalization, a navigation policy that learns to pick appropriate actions, and an external memory for leveraging past experiences.

\subsection{Data Augmentation}

\begin{figure}[t]
    \centering
    \includegraphics{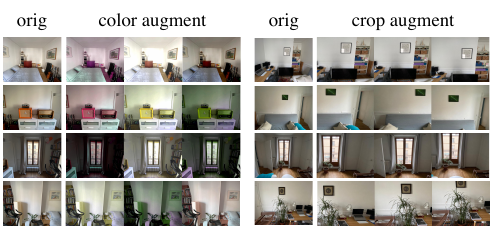}
    \caption{
      Illustration of data augmentation that we use to train our model.
      We consider both color jittering (left) and random crops (right).
      For a panoramic observation with $v$ views, the parameters of the augmentation are sampled independently.
    }
    \label{fig:data}
\end{figure}

To improve the generalization capacity of our agent to unseen environments, we apply random transformations on the observations of the simulator by using classic data augmentation techniques.
We use two kinds of data transformations: (i) \textbf{random cropping} that increases the input image size and takes a random crop of the original size in it, and (ii) \textbf{color jitter} that randomly changes the brightness, contrast, saturation and hue levels of the image.
An illustration of these transformations is shown in \autoref{fig:data}.
At navigation time, the agent receives the current and the goal observations from the simulator at each timestep. 
We apply both transformations sequentially to each of the $v$ views of the current and goal observations independently, producing $x_t$ and $x^*$ respectively.
This process allows for more visual diversity in the training data.

\subsection{Navigation Policy} 
Once the current and goal observations pass through the data augmentation phase, we use them in the navigation policy module, which computes the probability distribution over all possible actions: $\pi(a_t | x_t, x^*)$.
	 
First, the policy encodes each observation separately, as shown in \autoref{fig:model}.
We encode the current observation by feeding each of the $v$ views separately to the same convolutional neural network.
The $v$ vectors resulting from this operation are concatenated and passed into a fully-connected network.
This dimension-reduced output is then fed into a 2-layer Long Short-Term Memory (LSTM) along with a representation of previous actions, and the resulting vector $w^\text{obs}_t$ represents the embedding of the current observation at step $t$.

To encode the goal observation $x^*$, we process it through the same convolutional neural network as the current observation.
However, the outputs corresponding to the different views are added together instead of being concatenated, so as to make the representation of the goal rotation-invariant. 
We denote by $w^\text{goal}_t$ the resulting feature vector at step $t$.

Next, we make a joint representation by concatenating the current and goal feature vectors, before passing it through a fully-connected network to output an action $a_t$ as follows:
\begin{align}\label{eq:joint}
	w_t^\text{joint} &= \texttt{cat}(w^\text{obs}_t, w^\text{goal}_t), \\ \label{eq:out}
	\pi(a_t | x_t, x^*) &= \texttt{FC}(w_t^\text{joint}) .
\end{align}

\subsection{External Memory}
To add an external memory mechanism to the navigation policy, we first train a state-embedding network in a self-supervised fashion, as described in \autoref{sec:rnet-training}.
This network, trained to detect nearby locations, allows us to build an external memory containing representations of past observations. 
To leverage this memory, we add an attention module to the navigation policy. 

\subsubsection{Training a State-Embedding Network}
\label{sec:rnet-training}
Before learning the navigation policy, we train a state-embedding network to learn representations of the environment's locations.
The motivation for introducing this network is to encourage nearby locations in the environment to have similar representations, while ensuring distant locations to have different ones.
However, since we do not have access to the agent's position, the notion of distance between locations in the environment cannot be computed directly.
As in \cite{savinov2018episodic}, we will use the number of steps taken by an agent with a random policy to approximate this distance.

We let an agent with random policy explore the environment for T steps and denote by $(x_1, ..., x_T)$ the corresponding sequence of observations.\footnote{We ensure that the length of the computed sequence is $T$ by removing the \texttt{STOP} action from the action space.}
We then define a reachability label $y_{ij}$ for each pair of observations $(x_i, x_j)$ that depends on their distance in the sequence.
More precisely,
\begin{equation}
	y_{ij} = \begin{cases}
			1 \quad \text{if} \quad |i - j| \leq k, \\
			0 \quad \text{otherwise},
		 \end{cases}
	\quad \text{for } 1 \leq i, j \leq T 
\end{equation}
where $k$ is a hyperparameter.

We train a siamese neural network $R$, to predict the reachability label $y_{ij}$ from a pair of observations $(x_i, x_j)$. 
$R$ is defined by a convolutional network $g$ to embed the observations, and a fully-connected network $f$ to compare the embeddings, i.e.,
\begin{equation}
	R(x_i, x_j) = f(g(x_i), g(x_j)) .
\end{equation}
We apply the same data augmentation techniques to observations during this reachability network training phase.

\subsubsection{Episodic Memory}

Once we have the reachability network that can distinguish observations from nearby and distant locations, the agent can collect a compact memory of previously visited states.

We follow a process similar to \cite{savinov2018episodic} for building episodic memory. 
At timestep $t$, the agent has a memory buffer $M_{t-1}$ with embeddings from observations seen at previous timesteps.
Since storing every observation seen by the agent would be inefficient, we store only observations that are considered novel, i.e., distant from the current memory vectors.
In other words, at each timestep, we use the network $R$ to compute a reachability score between the current observation $x_t$ and the memory buffer $M_{t-1}$. 
This score corresponds to the maximum value obtained when comparing the current observation to every vector in the memory, i.e.,
\begin{equation}
	r(x_t, M_{t-1}) = \max \{ \ f(g(x_t), m) \ | \  m \in M_{t-1} \}.
\end{equation}

We then add the embedding of current observation to the memory if the reachability score is lower than a threshold, i.e., distant enough from the memory vectors. 
In other words,
\begin{equation}
	M_t = \begin{cases}
		M_{t-1} \cup g(x_t) \quad & \text{if} \; r(x_t, M_{t-1}) < \tau, \\
		M_{t-1} \quad 		    & \text{otherwise},
	      \end{cases}
\end{equation}
where $\tau$ is the \textit{reachability threshold} hyperparameter.
The episodic memory is reset after each episode.

This process for building episodic memory can be extended so that it persists across episodes within the same scene.
The impact of using such a cross-episodic memory is studied in \autoref{sec:ltm}

\subsubsection{Attention Module}
The navigation policy can leverage the episodic memory to move towards the target observation.
Rather than using an explicit planner on the memory, we take advantage of the information stored in them implicitly, using an \emph{attention module}. 

Our attention module has a multi-layer architecture similar to transformers~\cite{vaswani2017attention}.
Each layer consists of a multi-head attention sublayer (\texttt{Attn}), followed by a feedforward sublayer (\texttt{FF}).
See \cite{vaswani2017attention} for more details about these sublayers.
However, unlike transformers our attention module attends over a fixed set of vectors.
It comprises $N$ layers,
\begin{equation}
	\begin{aligned}
		z_t^{l} &= \texttt{FF}\left( \texttt{Attn}(z_t^{l-1}, M_t) \right), \quad \text{for } l \le N. \\
	\end{aligned}
\end{equation}
Here, $z_t^l$ is the output from the $l$-th layer, but the initial input $z_t^0$ is obtained by a linear transformation of the joint representation computed in \autoref{eq:joint}, $w_t^\text{joint}$.

The output from the attention module is then concatenated with the joint representation in \autoref{eq:out}, so the final action is now computed as:
\begin{equation}
	\pi(a_t | x_t, x^*) = \texttt{FC}(\texttt{cat}(w_t^\text{joint}, z_t^{N})) .
\end{equation}



\section{Experimental Results}

\begin{table*}[t]
  \centering
  \begin{tabular}{@{} l c ccc c cc c cc c cc c cc  @{}}
    \toprule
    &&  \multicolumn{3}{c}{Observation Type} && \multicolumn{2}{c}{Easy} && \multicolumn{2}{c}{Medium} && \multicolumn{2}{c}{Hard} && \multicolumn{2}{c}{Overall} \\ 
    \cmidrule{3-5}  \cmidrule{7-8} \cmidrule{10-11} \cmidrule{13-14} \cmidrule{16-17}
     Model && RGB & Pose & Depth && Succ & SPL && Succ & SPL && Succ & SPL && Succ & SPL \\
           \midrule
    Neural Topological SLAM [\textbf{NTS-D}]~\cite{chaplot2020neural} && {✗} & {✗} & {✗} && \textbf{0.87} & \textbf{0.65} && 0.58 & 0.38 && 0.43 & 0.26 && 0.63 & 0.43 \\
    \midrule
    ResNet + GRU + IL~\cite{chaplot2020neural} && \multirow{4}{*}{✗}& \multirow{4}{*}{✗}& \multirow{4}{*}{-}  && 0.57 & 0.23 && 0.14 & 0.06 && 0.04 & 0.02 && 0.25 & 0.10 \\
    Target-Driven RL~\cite{zhu2017target}       &&  & & && 0.56 & 0.22 && 0.17 & 0.06 && 0.06 & 0.02 && 0.26 & 0.10 \\
    Active Neural SLAM~\cite{chaplot2019learning} &&  & & && 0.63 & 0.45 && 0.31 & 0.18 && 0.12 & 0.07 && 0.35 & 0.23 \\
    Neural Topological SLAM [\textbf{NTS}]~\cite{chaplot2020neural} &&  & & && 0.80 &0.60 && 0.47 & 0.31 && 0.37 & 0.22 && 0.55 & 0.38 \\

	  \midrule
    SPTM~\cite{savinov2018semi} && {✗} &{-} & {-} &&0.64 & 0.35 && 0.52 & 0.27 && 0.36 & 0.19 && 0.51 & 0.27 \\ 
    \midrule
     \textbf{Ours} && {✗} &{-}&{-} && 0.78 & 0.63 &&  \textbf{0.70}& \textbf{0.57} && \textbf{0.60} & \textbf{0.48} && \textbf{0.69}&    \textbf{0.56}     \\ 
    \bottomrule
  \end{tabular}
  \caption{
    Comparison of our proposed model with several baselines and state-of-the-art approaches.
    The ``observation type" column shows the type of observation for each method: raw pixel observations (RGB), depth map (Depth), and position information (Pose).
    We report success rate and SPL, over three levels of difficulty.
    Our method establishes a new state of the art, e.g., doubling the SPL on \emph{hard} episodes, while not requiring any pose or depth information.
}
  \label{tab:main-result}
\end{table*}

\subsection{Implementation Details}
\label{impl_details}

\parag{Task Setup}
We conducted all of our experiments on the Habitat~\cite{savva2019habitat} simulator with the Gibson~\cite{xia2018gibson} dataset, which contains a set of visually-realistic indoor scenes.
We used the standard 72/14 train/test scene split for this dataset.
As stated earlier, \textbf{we do not use the agent's pose or depth sensor information}.
The forward step range and turn angle are set to their standard values $(0.25\text{m}, 10^\circ)$ for navigation episodes and $(1\text{m}, 30^\circ)$ when training the reachability network.
The maximum number of steps in an episode is 500, and the success distance $l$ is $1\text{m}$.
In addition to the success rate, we also use success weighted by inverse path length (SPL)~\cite{anderson2018evaluation} as an evaluation metric.
SPL takes into account the length of the path that the agent has taken to the goal.

\parag{Training the Reachability Network}
We generate one trajectory per train scene from an agent with a random policy.
We allow 5k steps for each trajectory and remove the stop action from the action space.
This results in a total of $360\text{k}$ steps from 72 scenes to train the reachability network.
From each trajectory, we sample 1k positive pairs (within $10$ timesteps) and 1k negative pairs, yielding a dataset of $144\text{k}$ image pairs.
We implement the reachability network as a siamese network with ResNet18 for the embedding function $g$. 
Each of the $v$ views from the RGB observation is passed through the ResNet separately. 
We sum the resulting outputs to form the embedding vector of a panoramic observation.
The comparison function $f$ is composed of two hidden layers of dimension 512 with ReLU activations.
We train this network using SGD for 30 epochs with a batch size of 256, a learning rate of 0.01, a momentum of 0.9, a weight decay of $1\mathrm{e}-7$, and no dropout.

\parag{Training Data for the Navigation Policy}
We generated 9k navigation episodes in each training scene, following the protocol of \cite{chaplot2020neural}.
\footnote{
  We obtained the generation procedure from the authors of~\cite{chaplot2020neural} by e-mail. 
  The dataset is available at: \url{https://github.com/facebookresearch/image-goal-nav-dataset}
}
We split our navigation episodes into three levels of difficulty, based on the distance between the start and the goal locations: \emph{easy} (1.5 - 3m), \emph{medium} (3 - 5m), and \emph{hard} (5 - 10m).
For each scene, we sample 3k start-goal location pairs per level of difficulty, resulting in 648k train episodes.
Similarly, we sample 100 episodes per test scene and per level of difficulty, resulting in 4.2k test episodes.

\parag{Navigation Policy Implementation}
At the beginning of each episode, the simulator generates the observation from the goal location as a $v \times 3 \times 128 \times 128$ panoramic RGB image and gives it to the agent as target observation.
We used ResNet18 with shared weights for encoding the current and target observations in the policy.
The size of the embedding space is 512.
We concatenate the encoder's outputs for the $v$ views of the observation and feed them into an LSTM with two recurrent layers.
Our attention module consists of 4 stacked layers of a 4-headed attention network.
We set the buffer's capacity to 20 for the episodic memory.
We train the policy using DDPPO~\cite{wijmans2019dd} for 50k updates, with 2 PPO epochs, a forward of 64 steps, an entropy coefficient of $0.01$, and a clipping of $0.2$.
We used the Adam optimizer with a learning rate of $9\mathrm{e}-5$.

\parag{Data Augmentation}
For the training stages of both the reachability network and the navigation policy, we used random cropping with a minimum scale of 0.8 and color jittering with value 0.2 for brightness, contrast, saturation, and hue levels.
These transformations are applied at two different levels when training the navigation policy: (i) when the agent samples the action $a_t$ from the policy, and (ii) during the forward-backward in PPO.
Note that, for these two steps, the transformation applied to the images is independent and results in different input images.


\subsection{Comparison with the state of the art}
We consider several methods in this comparison: an adaptation of SPTM~\cite{savinov2018semi} to Habitat, as well as four approaches taken from~\cite{chaplot2020neural}.

\noindent
\textbf{ResNet + GRU + IL}~\cite{chaplot2020neural}: A simple baseline consisting of ResNet18 image encoder and a GRU-based policy trained with imitation learning.

\noindent
\textbf{Target-Driven RL}~\cite{zhu2017target}: A vanilla baseline trained with reinforcement learning, similar to our ablated variant.

\noindent
\textbf{Active Neural SLAM}~\cite{chaplot2019learning}: An exploration model based on metric maps, adapted to navigation.

\noindent
\textbf{Neural Topological SLAM (NTS)}~\cite{chaplot2020neural}: The previous state-of-the-art method for navigation in unseen environments, based on neural SLAM and endowed with a topological graph for long-term planning.
We compare against two versions of this method: \textbf{NTS}, that uses as input the RGB observation and the pose sensor, and \textbf{NTS-D} that uses the depth map in addition to the two other modalities.

\noindent
\textbf{SPTM}~\cite{savinov2018semi}: A navigation model based on explicit construction of a graph over past experience.
The method requires an exploration episode given \emph{a priori}.
To run this baseline, we adapt the available code\footnote{\url{https://github.com/nsavinov/SPTM}} to Habitat. 
We train the reachability and locomotion networks on trajectories from the training set. 
For evaluation we used a random exploration trajectory of 10k steps, and build a graph with 100k top-scoring edges.
As this method does not implement a \texttt{STOP} action, we give it advantage and automatically end the episode when the agent gets within 1m of the goal.

The first four methods are designed for episodic navigation, i.e. the validation episodes are performed on completely unseen scenes, without pre-exploration.
Conversely, \textbf{SPTM} is intended for setups where a pre-exploration phase is allowed on the test scenes before starting the actual navigation phase, and therefore uses more privileged information.

We trained our model on the \emph{easy}, \emph{medium}, \emph{hard} combined dataset for 500M steps for three random seeds, and evaluated it on the corresponding test set.
As shown in \autoref{tab:main-result}, the performance obtained with our memory-augmented model is superior to that of previous work by a significant margin: $+13\%$ SPL on average compared to \textbf{NTS-D} and $+29\%$ SPL vs.\ \textbf{SPTM}.
We obtain this strong performance while, unlike \textbf{NTS-D}, not using position nor depth information.
We note that the success rate on \emph{easy} episodes is lower for our method than \textbf{NTS-D} ($-9\%$).
This is potentially due to the lack of position and depth features or the use of discretized panoramic observations, both of which are particularly useful for nearby goals.
Note that the four methods from~\cite{chaplot2020neural} require 25M steps of training until convergence in comparison to our 500M steps.
This difference in training horizon is due to several factors, including the absence of position information in the observations for our setup, and the absence of an explicit planner in our method.
Rather than comparing all the methods at a fixed-step budget, when they may not be trained optimally, we made the fair choice to compare all the methods after the convergence of their respective optimization processes.


\begin{figure}[t]
  \centering
  \includegraphics[width=\linewidth]{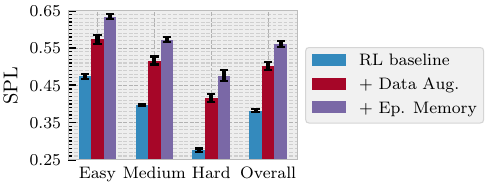}
  \caption{Ablation study. 
  We present the SPL for three variants of our model.
  All models are trained for 500M steps on a combination of \emph{easy}, \emph{medium}, and \emph{hard} episodes.
  }
  \label{fig:ablation}
\end{figure}

\subsection{Ablation Study and Analysis}
\label{sec:ablation}

We perform an ablation study to empirically validate the design choices of our model.
We evaluate the performance of the simple target-driven RL baseline, as well as the improvements brought by the data augmentation module and the episodic memory. 
To this end, we train three variants of our model: (i)~\textbf{RL baseline}, the vanilla target-driven RL baseline to which we consecutively add (ii)~\textbf{Data~Aug.}, the data augmentation module, and (iii)~\textbf{Ep.~Memory}, the episodic memory module.
We train all models on this dataset for 500M steps for three random seeds and report the average SPL for each level of difficulty in \autoref{fig:ablation}.

First, we see that using data augmentation when training a RL-based navigation policy in this context improves SPL significantly: the gap with the vanilla baseline is $+12\%$ \emph{overall}.
Second, we observe that the episodic memory-based policy improve over the very competitive data-augmented baseline ($+6\%$ \emph{overall}).
Finally, we note that the data-augmented baseline significantly outperforms the state of the art method \textbf{NTS} ($+12\%$ \emph{overall}), highlighting the power of this simple end-to-end model.

\begin{table}[h]
  \centering
  \begin{tabular}{@{}lccccc@{}}
  \toprule
	  Model           	      & Easy& Med. & Hard & Extra & Overall \\ 
  \midrule
   Ep.~Memory & 0.560 & 0.505 & 0.428 & 0.138 & 0.407      \\
   LT~Memory     & \textbf{0.569} & \textbf{0.528} & \textbf{0.455} & \textbf{0.161} & \textbf{0.428}  \\ 
  \bottomrule
  \end{tabular}
  \caption{
    Evaluating the impact of a memory that persists across episode.
    We train both models for 500M steps on a combination of \emph{easy,} \emph{medium}, \emph{hard}, \emph{extra} episodes.
    The long-term memory improves the performance of the model, especially on the \emph{hard} and \emph{extra} splits.
  }
  \label{tab:ltmem}
\end{table}

\subsection{Impact of a Long-Term Memory}
\label{sec:ltm}

We also studied the impact of a memory that persists across episodes in the same scene, motivated by the observation that embodied agents, once deployed, do not simply cease to exist after an episode has ended: they persist, and so should their memories.
We therefore added an additional memory to the episodic one, that remains for $100$ episodes in the same scene.
We compare this model with long-term memory (\textbf{LT~Memory}) to the one with episodic memory only (\textbf{Ep.~Memory}) in \autoref{tab:ltmem}.
For this experiment, we generated an additional split to the \emph{easy}, \emph{medium}, and \emph{hard} episodes, named ``\emph{extra},'' for which the distance between start and goal locations is 10~-~15m.
This dataset, voluntarily more challenging, is intended to highlight the importance of a memory that persists across episodes.
The long-term memory improves the performance of the model over the episodic one ($+2.1\%$ \emph{overall}), and the difference is more important on \emph{hard} and \emph{extra} episodes ($+2.7\%$ and $+2.3\%$ respectively) than on \emph{easy} ones ($+0.9\%$).

\begin{figure*}[t]
	   \centering
	   \includegraphics{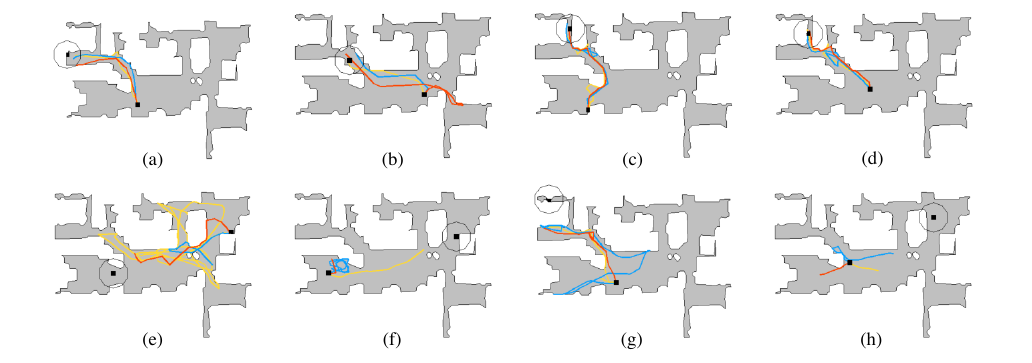}
	   \caption{
	       The top (resp. bottom) row shows trajectories from test episodes with the highest (resp. lowest) SPL on the Eastville scene.
	       Start and goal locations are shown in black, with the goal being circled by a line showing the success area.
	       Results are shown for our model with memory trained on the standard dataset, for 3 seeds (corresponding to the 3 colors) on the \emph{hard} validation split.
	   }
	   \label{fig:traj}
\end{figure*}

\begin{table}[h]
  \centering
  \begin{tabular}{@{}l cccc@{}}
    \toprule
	  Number of views ($v$)  & 1 & 3 & 4 & 6 \\
    \midrule
    SPL & 0.08 & 0.31 & 0.36 & 0.36 \\
    Frames per sec. & 1890 & 2000 & 2080 & 2340 \\
    \bottomrule
  \end{tabular}
  \caption{
    Analysis of SPL obtained with the \textbf{RL baseline} for various panoramic view configurations.
    We report average SPL and the number of frames processed per second for each configuration.
   }
  \label{tab:rotation_numbers}
\end{table}

\subsection{Analysis and Qualitative Visualisations}

\parag{Panoramic Observations}
As described in \autoref{sec:problem}, we generate panoramic observations by equally spaced planar observations around the agent.
In this experiment, we compare the performance of the vanilla \textbf{RL baseline} model trained with $1$, $3$, $4$ and $6$ views around the agent.
We let the model train for 500M steps for three random seeds and report the average SPL obtained by this agent in \autoref{tab:rotation_numbers}.

We note that an agent trained with a single view, i.e., without panoramic observations, completely fails to learn a successful policy, obtaining only $0.08$ SPL.
This result is quite intuitive, as the relative localization with respect to the goal is made easier by multiple views.
Better performance is obtained with either four or six views, with an SPL of $0.36$.
We also see that there is a tradeoff between performance and additional runtime required with more views.
Thus, we run all the variants of the models with four views.

\parag{Data Augmentation and Overfitting}
We investigate how data augmentation and adding more training scenes allow us to bridge the train~/~test gap observed when navigating an unseen environment.
To this end, we generate an extended training set by considering 150 additional scans, which are usually rated as being of poor quality.
We perform this experiment for the vanilla \textbf{RL Baseline}, as well as its data-augmented version trained on the standard dataset (\textbf{Data Aug.}), and the extended one (\textbf{More Scenes}).

\begin{figure}[h]
  \centering
  \includegraphics[width=\linewidth]{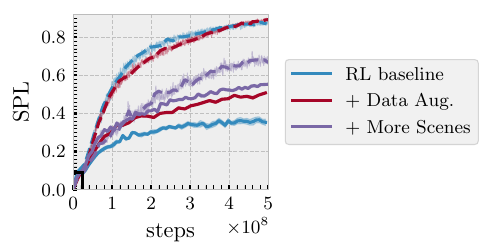}
  \caption{
	  Performance measured in SPL as a function of training steps taken in the environment for the \textbf{RL baseline}, with data augmentation (+ Data Aug.) and by adding more training scenes (+ More Scenes).
    We report both the training (dashed line) and test (solid line) SPL on the same figure.
    The generalization gap is large and can be reduced by using data augmentation.
    It can be reduced significantly further by training the model on a larger set of scenes.
	The black mark at 25M steps corresponds to the \textbf{Target-Driven RL} baseline as reported in~\cite{chaplot2020neural}.
  }
  \label{fig:insta}
\end{figure}

As seen from the train and test SPL for the three methods in \autoref{fig:insta}, the generalization gap is large (almost $65\%$ for the \textbf{RL baseline}).
The use of data augmentation allows to remedy this problem reducing this gap to about $40\%$.
Training the model on more scenes reduces this discrepancy even further to only $15\%$.
We also observe that data augmentation not only improves the test performance but also helps faster convergence ($+2\%$ on the train SPL), as opposed to what is usually observed in supervised learning.

Notice that in the figure, we indicate the 25M step mark~--~this is the number of environment steps used to train the \textbf{Target-Driven RL} baseline reported in~\cite{chaplot2020neural}.
We see that the performance of a reinforcement-based model with such a limited number of steps is quite poor, and that convergence is far from being achieved at this point.
Though sample efficiency is an important challenge in robotics applications, the \textbf{RL baseline} should be trained sufficiently to leverage its full potential.
Indeed, after 500M steps of training, it reaches 0.37 SPL \emph{overall}, which is almost on par ($-1\%$) with the performance of the method proposed in~\cite{chaplot2020neural}.

\parag{Qualitative Results} \autoref{fig:traj} shows example success and failure cases from episodes of the test dataset.
We see that our agent successfully learns to navigate to challenging locations, which are distant from the start location~(\ref{fig:traj}-a) and/or located at extremities of the scenes~(\ref{fig:traj}-c, \ref{fig:traj}-d). 
Moreover, our agent shows interesting skills, like bypassing obstacles~(\ref{fig:traj}-c) or looking around in a room~(\ref{fig:traj}-b).
From the failure cases (bottom row in the figure), we see that our agent has some undesired behaviour.
For example, it can get stuck in a loop~(\ref{fig:traj}-f), stop too early ~(\ref{fig:traj}-h), or fail to reach some extremely challenging goals~(\ref{fig:traj}-g).

\section{Conclusion}
In this paper, we have presented a memory-endowed agent that we train end-to-end with reinforcement learning.
This memory is accessed in the navigation policy using a transformer-inspired neural network with attention modules.
We evaluated our agent on the challenging task of image-goal navigation, and have shown that it surpasses previous work by a large margin.
This impressive performance is obtained from RGB observations only, i.e., without using any position information, though it requires more training steps.
In future work, we plan to improve the training of the reachability network and make the policy better exploit the memory.

\section*{Acknowledgements.}
Karteek Alahari is supported in part by the ANR grant AVENUE (ANR-18-CE23-0011).

\bibliographystyle{IEEEtran}
\bibliography{refs}

\appendix

\section{Analysis of the Reachability Network}
\label{sec:reach_analysis}
We begin with additional details about training the Reachability Network and then present a few qualitative visualisations.

\begin{figure}[h]
  \centering
  \includegraphics[width=\linewidth]{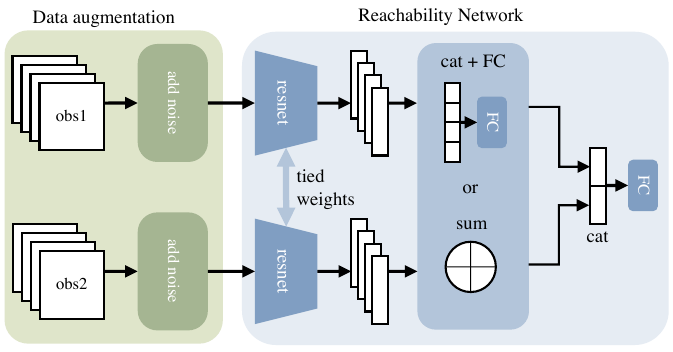}
  \caption{Architecture of the Reachability Network. We adapted the architecure from~\cite{savinov2018episodic} by using a data augmentation module and a layer that handles panoramic observations. 
  The output of the last fully-connected module is the similarity score between the two observations.}
   \label{fig:rnet}
\end{figure}

\subsection{Architecure \& Training Details}

We adapted the architecture of the Reachability Network presented in~\cite{savinov2018episodic} to work with panoramic views in realistic environments. 
An illustration of this model is shown in Figure~\ref{fig:rnet}.
The main differences with~\cite{savinov2018episodic} are the use of a data augmentation module, that applies transformations to the RGB input of the network, and an additional layer to handle panoramic observations.
This layer aggregates the ResNet output for each of the $v$ views into one feature vector.
We experimented with two architectures for this layer: (i) \textbf{cat + FC}: where we concatenate the ResNet output for each of the $v$ views and feed this large vector into a one-layer fully-connected module, and (ii) \textbf{sum}: where we simply sum these $v$ vectors.
Contrary to (i), (ii) has the interesting property of yielding embeddings that are rotation invariant---the resulting feature vector from a location of the environment will be the same for every direction of the agent.

\begin{figure}[h]
  \centering
  \includegraphics[width=\linewidth]{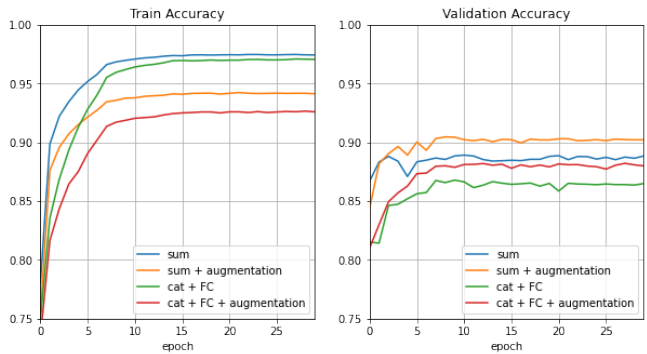}
  \caption{Training and validation curves for the Reachability Network. We tested four setups, that compare the use of data augmentation and the choice of the aggregation layer for panoramic features.}
   \label{fig:performance}
\end{figure}

We compare the train and validation performance of these two design variants, and the influence of using data augmentation for training the Reachability Network. 
The results are shown in Figure~\ref{fig:performance}.
First, we observe that using data augmentation reduces overfitting in both the setups, and yields a better validation accuracy in these two cases.
Second, we see that summing panoramic features allows to achieve better train and validation performance than concatenating them. 
One explanation for this is that the rotation invariance explained above is facilitating the learning.
For all of our navigation policy learning experiments, we chose the setup \textbf{sum + augmentation} for training the Reachability Network that encodes the memory vectors.

\begin{figure}[h]
\centering
   \includegraphics{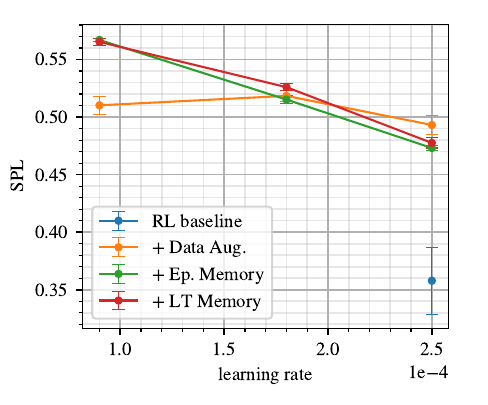} \\
   (a)\\
   \includegraphics{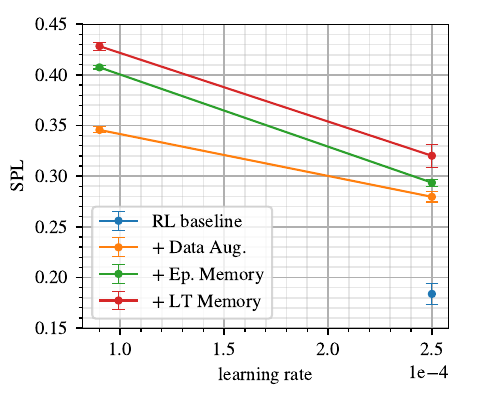}\\
   (b)
\caption{Overall SPL for several values of the learning rate.
  We plot the results for the four ablated variant of our method, when trained on the standard dataset (a) and on the extra one (b).
  Each experiment was ran for three random seeds.
  Methods with external memory benefit from a lower learning rate.}
 \label{fig:lr_plot}
\end{figure}

\subsection{Qualitative Visualisations}

\begin{figure*}
	   \centering
	   \begin{tabular}{ccc}
		   \includegraphics[width=5.5cm]{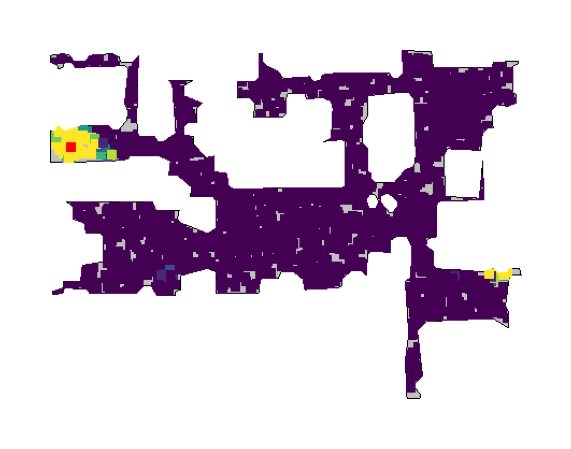}&
		   \includegraphics[width=5.5cm]{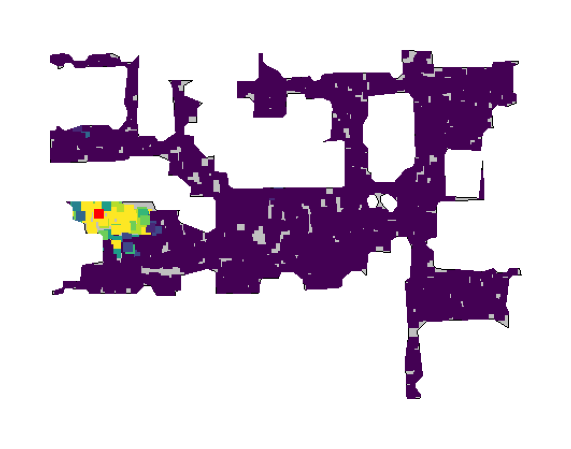}&
		   \includegraphics[width=5.5cm]{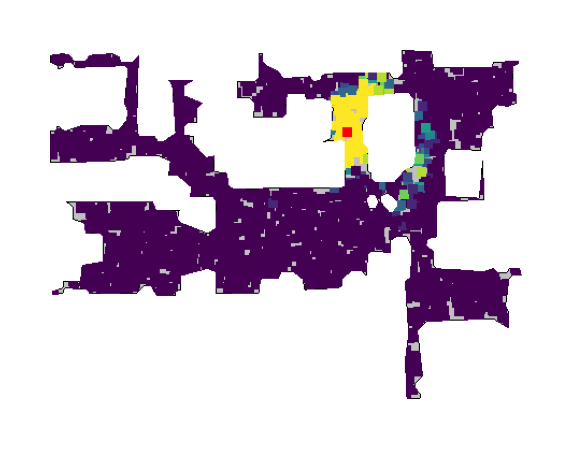}\\
		    (a) & (b) & (c) \\ 
		   \includegraphics[width=5.5cm]{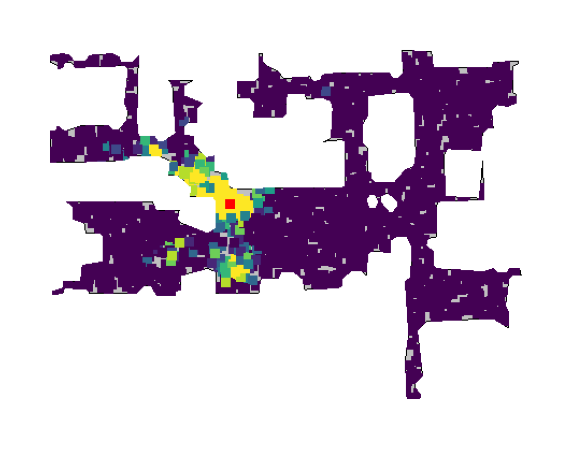}&
		   \includegraphics[width=5.5cm]{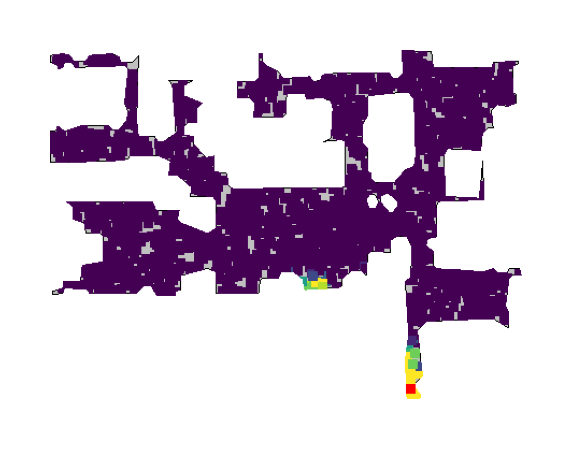}&
		   \includegraphics[width=5.5cm]{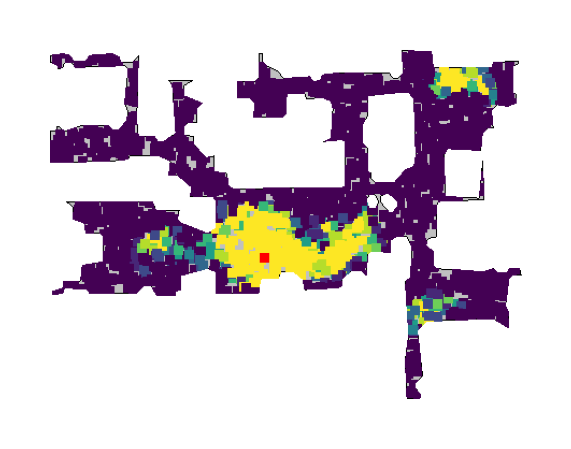}\\
		  (d) & (e) & (f) \\
	   \end{tabular}
	      \caption{Heat-maps of the similarity score between the observation (the red point) and the observations at $N = 2000$ points sampled randomly in the environment. 
	      The colour at a location corresponds to the similarity score at that location: low values, close to 0, are in dark violet and high values, close to 1, in yellow.
	      These visualisations were performed on Eastville: an environment of the validation set.}
	           \label{fig:rnet-values} 
\end{figure*}

We visualise the quality of the Reachability Network with the following experiment.
First, we put the agent at a random location in the environment and sample an observation $x$ from there. 
Then, we randomly sample $N$ observations in the environment and for each of these observations, we compute their similarity score with observation $x$, using the Reachability Network.
We present these results on a heat-map, where the colour at a location represents the corresponding similarity score.
Some examples are shown in Figure~\ref{fig:rnet-values}.
We see that the high similarity scores are at locations that are around the comparison observation, which implies that the Reachability Network performs well at learning representations that are similar for nearby locations, and dissimilar for representations that are far away.
Since these experiments are shown on a validation environment, we note that the Reachability Network generalizes well to unseen environments.

\begin{figure*}
	\centering
	\begin{tabular}{c}	
		   \includegraphics[width=\textwidth]{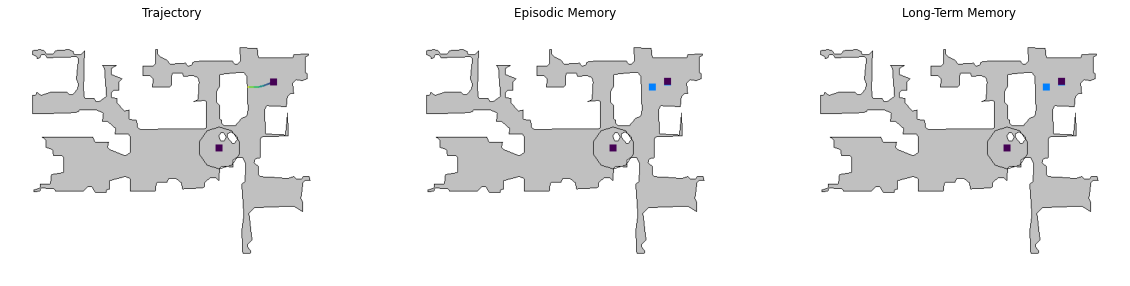} \\
		   (a) Step 1 \\
		   \includegraphics[width=\textwidth]{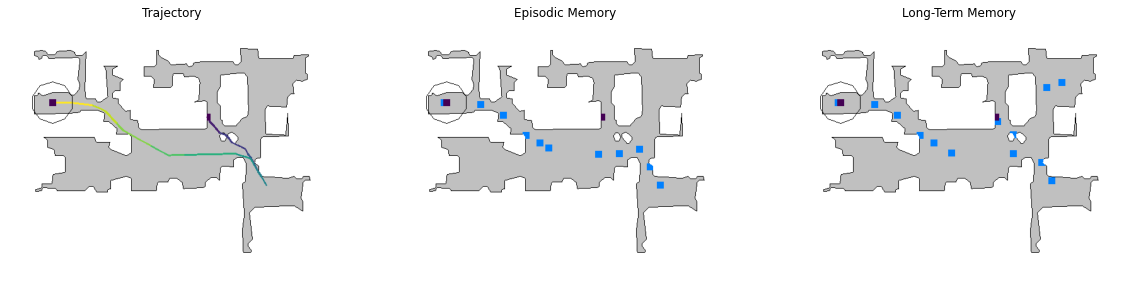} \\
		   (b) Step 2 \\
		   \includegraphics[width=\textwidth]{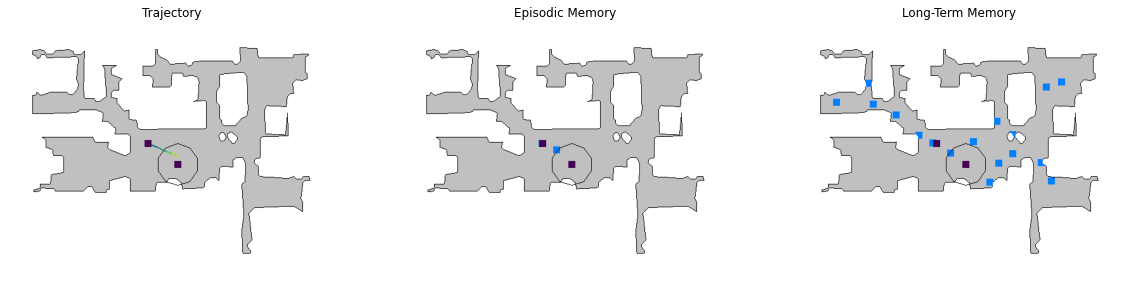} \\
		(c) Step 5 \\
		   \includegraphics[width=\textwidth]{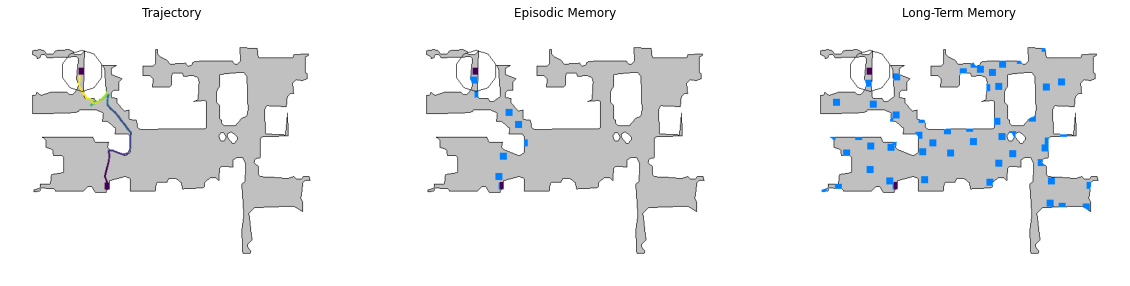}\\
		   (d) Step 100 \\
	\end{tabular}
	    \caption{Visualisation of the agent's trajectory, episodic and long-term memories for first, second, fifth and 100th episode in the Eastville environment.
	    The start and goal locations are shown in black, goal location being circled by a line showing the success area. 
	    The blue points represent the location of the episodic and long-term memory vectors.
	    The episodic memory is reset after each episode, while the long-term memory remains for 100 episodes in the same scene.}
  	     \label{fig:traj-em-ltm}
\end{figure*}

We also visualise the state of the episodic and long-term memories for consecutive validation episodes in Figure~\ref{fig:traj-em-ltm}.
From this, we observe how these memories are filled through consecutive validation episodes.
After 100 navigation episodes (\ref{fig:traj-em-ltm}-d), the long-term memory is well filled and covers most of the environment.
This allows the agent to reach challenging goals.

\section{Optimizing the learning rate}
\label{sec:results}

Learning rate tuning is an important step for training our models.
Figure~\ref{fig:lr_plot} shows the \emph{overall} SPL for the four ablated variants of our method and several values of the learning rate when trained on the standard dataset (~\ref{fig:lr_plot}-a) and on the \emph{extra} one (~\ref{fig:lr_plot}-b).
Each method was trained for three random seeds.
We observe that the methods that contain external memory benefit a lot from decreasing the learning rate, as opposed to the data-augmented baseline.

\section{\textbf{SPTM}~\cite{savinov2018semi} implementation details}
\label{sec:sptm}

\begin{figure}[b]
  \centering
  \includegraphics[width=0.7\linewidth]{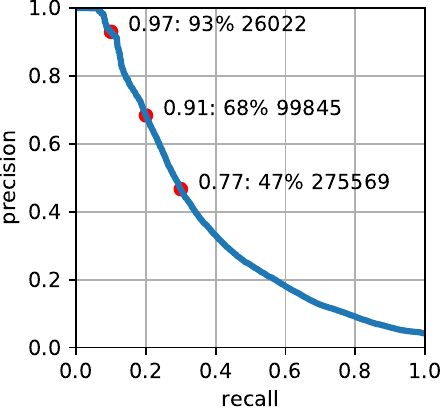}
  \caption{
    PR curve for edge prediction for \textbf{SPTM}~\cite{savinov2018semi}.
    We choose the number of edges to keep for navigation based on this evaluation.
    In our experiments, we kept the 100k top scoring edges.
  }
   \label{fig:pr-sptm}
\end{figure}

As mentionned in the main paper, we compare our method to SPTM.
In order to compare to this method, we need to adapt the code to the Habitat~\cite{savva2019habitat} environment.
We use RGB observations of size $160 \times 120$ pixels. 
We train the Reachability and Locomotion networks with a learning rate of 1e-4 for 300 ``epochs'', with a batch size of 64.
To evaluate it, we use an exploration trajectory obtained with a random policy rolled-out for 10k steps. 

In order to select the number of edges to infer, we look at the precision and recall of the Reachability Network trained in this context.
We label all edges as a positive if it connects two nodes that are physically less than 1m apart. 
Given this label, and the scoring obtained from the Reachability Network, we plot the precision-recall curve that we present in Fig.~\ref{fig:pr-sptm}.

On this plot, we see that if we set the threshold for the Reachability Network to $0.9$, we obtain a precision of $68\%$ for a recall of $20\%$.
In that case we pick $99845$ edges to be added to the graph.
This precision and recall tradeoff leads to good performance, so we decided to keep 100k top scoring edges in our experiments.

\end{document}